\documentclass[authoryear,5p,times,nopreprintline]{elsarticle}

\usepackage{amssymb}
\usepackage{amsmath}
\usepackage{bm}
\usepackage{mathtools}
\usepackage{graphicx}
\graphicspath{{./}}
\usepackage{microtype}
\usepackage{booktabs}
\usepackage{multirow}
\usepackage{makecell}
\usepackage{array}
\usepackage{siunitx}
\usepackage{subcaption}
\usepackage{tabularx}
\usepackage{float}
\usepackage{xcolor}
\usepackage{tikz}
\usetikzlibrary{arrows.meta,calc}
\usepackage{etoolbox}
\usepackage{textcomp}
\usepackage{url}
\usepackage{hyperref}
\hypersetup{
  hidelinks,
  pdftitle={Coding-agents can replicate scientific machine learning papers},
  pdfauthor={Atharva Hans, Ilias Bilionis}
}
\makeatletter
\renewenvironment{equation}%
  {\@beginparpenalty\predisplaypenalty
   \@endparpenalty\postdisplaypenalty
   \refstepcounter{equation}%
   \trivlist\item[]\leavevmode
   \hb@xt@\linewidth\bgroup\hfil$\m@th\displaystyle}%
  {$\hfil\llap{\@eqnnum}\egroup
   \endtrivlist}
\makeatother

\makeatletter
\patchcmd{\@@author}
  {}
  {}
  {}{}
\makeatother

\journal{Computer Physics Communications}

\begin{document}

\begin{frontmatter}

\title{Coding-agents can replicate scientific machine learning papers}

\author[aff1]{Atharva Hans\corref{cor1}\fnref{present}}
\ead{atharva.hans@lilly.com}

\author[aff1]{Ilias Bilionis}

\cortext[cor1]{Corresponding author.}
\fntext[present]{Present address: Eli Lilly and Company, Indianapolis, IN 46285, USA}

\affiliation[aff1]{
  organization={School of Mechanical Engineering, Purdue University},
  city={West Lafayette},
  state={IN},
  postcode={47907},
  country={USA}
}

\begin{abstract}
Scientific machine learning papers typically make computational claims, e.g., that the relative mean square error is less than 5\% or that the 95\% predictive credible interval covers the test data.
A coding agent can be prompted to replicate those claims from paper materials alone, but the prompt does not by itself reliably preserve progress or check whether generated evidence supports the paper's claims.
We introduce \emph{Paper-replication}, a workflow that makes each selected paper claim a target with recorded evidence, and implement it as a coding-agent skill.
The workflow makes the agent record those targets, reconstruct the paper's method, run computational experiments, link generated outputs to provenance and comparisons with the paper's claims, record where matched evidence appears in the replication report, and pass validation checks before completion.
We evaluate Paper-replication on twelve independent runs across four scientific machine learning papers.
All twelve workspaces pass the completion gate, and all \(158\) recorded targets are matched with report coverage.
Even in this completed workspace state, repeated runs differ in how papers are divided into targets, in numerical fidelity to the source papers, in elapsed replication time, in the number of intermediate executions replaced before final evidence is accepted, and in the rules used to accept evidence.
Paper-replication makes completion depend on workspace evidence and validation checks rather than on the agent's final message.
\end{abstract}

\begin{keyword}
Coding agents \sep Harness engineering \sep Paper replication \sep Scientific reproducibility \sep Scientific machine learning
\end{keyword}

\end{frontmatter}

\section{Introduction}\label{sec:intro}
Computational claims of scientific machine learning papers depend on differentiable solvers and neural differential equations~\citep{chen2018neural,rackauckas2020universal,hans2023stochastic,hans2024bayesian}, physics-informed neural networks and physics-informed learning objectives~\citep{raissi2019physics,karniadakis2021physics}, sparse discovery of dynamical systems and partial differential equations~\citep{brunton2016discovering,rudy2017data}, and Bayesian or sampling-based inverse-problem methods~\citep{stuart2010inverse,hans2020quantifying,hans2023bayesian,alberts2023physics}.
A reader can inspect the equations and algorithms in the text, but the reported results also depend on code, data, numerical solvers, preprocessing, stochastic training, and implementation choices~\citep{hans2024scalable,hans2025smurf}.
Reproducible-research practice therefore treats code, data, computational environments, and the record of how each result was produced as part of the scientific record~\citep{peng2011reproducible,stodden2016enhancing}.
It also emphasizes versioned project organization~\citep{wilson2017good}, explicit tracking of random seeds and intermediate results~\citep{sandve2013ten}, executable notebooks~\citep{kluyvertextordfeminine12016jupyter}, and workflow provenance~\citep{davidson2008provenance,belhajjame2013prov,goecks2010galaxy}.

Terminology around reproducibility and replicability varies across fields~\citep{barba2018terminologies}.
We define paper replication as the task of reconstructing a paper's method from paper materials, regenerating the computational results that support its claims, and recording evidence for each claim.
Replicating a paper from paper materials alone is a narrower and harder task than rerunning a released package.
The available materials may include the \LaTeX{} source, figures, tables, appendices, and data references, but they need not include sampled training sets, seeds, optimizer states, sampler states, preprocessing conventions, or plotting rules.
In machine learning, reproducibility taxonomies treat a paper-only record as weaker evidence than a runnable package with code, data, dependencies, and environment information~\citep{tatman2018practical}.
Surveys of artificial-intelligence papers also find that empirical details are often underreported~\citep{gundersen2018state}.
Stochastic training makes these omissions matter because data sampling, initialization, implementation choices, and hyperparameters can change the result even when the algorithmic description stays fixed~\citep{henderson2018deep,bouthillier2021accounting,pineau2021improving}.

Research code generation methods address one part of paper replication: they convert paper descriptions into code or implementation tasks.
PaperCoder is a multi-agent framework that turns machine learning papers into code repositories through planning, implementation analysis, and modular code generation~\citep{seo2025paper2code}.
ResearchCodeBench converts recent machine learning papers into executable implementation challenges~\citep{hua2026researchcodebench}, and ResearchCodeAgent uses a multi-agent system with planning, memory, and actions to codify methods from the machine learning literature~\citep{gandhi2025researchcodeagent}.
Other benchmarks evaluate algorithmic reproduction from natural-language-processing papers~\citep{xiang2025scireplicate}, reconstruction of masked language-modeling research code~\citep{yan2025lmr}, and progressive code-masking experiments~\citep{kim2025reproduction}.
These studies show that agents can help read papers, recover implementation details, and write or repair research code.

Paper-reproduction benchmarks shift the focus from code generation to reproducing reported results.
PaperBench asks agents to replicate papers from the International Conference on Machine Learning by reading the paper, building a codebase, and executing experiments, and grades attempts with author-informed rubrics~\citep{starace2025paperbench}.
CORE-Bench evaluates computational reproducibility when code and data packages are available~\citep{siegel2024core}.
Social-science benchmarks and systems ask agents to assess reproducibility from a paper and reproduction package~\citep{hu2025repro}, replicate claims across stages that include data retrieval, experiment design, execution, and interpretation~\citep{nguyen2026replicatorbench}, reproduce findings with original data and code~\citep{alizadeh2026ai}, repair controlled reproducibility failures~\citep{shah2026automating}, or reimplement analyses from a paper's method description and original data without the original code~\citep{kohler2026read}.
Recent benchmarks also extend the question to astrophysics~\citep{ye2025replicationbench}, materials science~\citep{huang2026can}, and collider physics~\citep{faroughy2026collider}.
These studies show that agents can run experiments and assess reproduced outputs.
They also show that paper replication is not only a question of whether an agent can produce code or run an experiment.
In many of these settings, a benchmark harness, a masked codebase, a reproduction package, an original dataset, or an author-informed rubric defines what the agent must produce and how the result is judged.
When replication starts from paper materials alone, the workflow must define this evidentiary standard.
The workflow must first identify the paper claims selected for replication as targets.
Each accepted target must then link its claim to the agent's reconstruction of the paper's method, a successful run, provenance, comparison evidence, and report coverage.
Without those links, generated output can look plausible even if it comes from a copied paper-provided asset, a substitute method, or an unrecorded run.
Existing work does not define paper replication as a target-level evidence contract in a persistent workspace.
It also leaves open how to make completion a workspace state rather than an agent final-message claim.

Coding agents provide the capabilities needed to carry out this workflow.
Codex~\citep{openai2026codexweb} and Claude Code~\citep{anthropic2026claudecode} can inspect files, edit code, run commands, and iterate on failures.
Coding-agent benchmarks test related capabilities by requiring agents to inspect repositories, edit code, run tests, and resolve issues~\citep{jimenez2024swe,yang2024swe}.
For paper replication, these capabilities map to inspecting the paper source, writing implementations, executing experiments, comparing outputs, and preparing a replication report.
These actions are necessary, but they do not by themselves define what counts as acceptable evidence for a replicated claim.

A prompt can tell the agent what standard to follow, but a prompt alone does not make that standard durable.
The agent must remember which targets remain, preserve assumptions across interruptions, distinguish generated outputs from paper-provided assets, and decide whether a comparison supports a claim.
Agent benchmarks report failures in long-horizon reasoning, decision making, and instruction following~\citep{liu2024agentbench}.
Studies of intrinsic self-correction also show that a language model cannot reliably judge and repair its own reasoning without external feedback~\citep{huang2024large}.
A prompt that asks the agent to finish the replication can then lead it to report completion before the workspace contains the required target-level evidence~\citep{manheim2018categorizing}.
For paper replication, the agent's final message is not enough evidence.
The workflow must store state and check evidence outside that message.

Paper replication therefore requires a workflow that leverages the coding agent's ability to work with files and run commands, while defining progress and completion through workspace records and validation checks.
A coding-agent skill provides one way to give this workflow to an existing agent~\citep{anthropic2025agentskills,openai2026codexskills}.
The skill is a reusable set of task instructions and workspace utilities for a class of tasks.
For paper replication, the skill can specify how the agent initializes or reopens the replication workspace, which records it writes, which checks it runs, and when a target can be marked as matched.

We introduce \emph{Paper-replication} as this workflow and implement it as a coding-agent skill.
Paper-replication uses targets as the unit for recording evidence and checking completion.
It addresses the prompt-only failure modes above through harness engineering~\citep{lopopolo2026harness}: it changes the environment in which the agent works by adding a persistent replication workspace and validation checks rather than relying on prompt instructions alone.
Starting from the paper materials, the agent records the target set in a reproduction matrix, keeps one active target in a task ledger, records its reconstruction of the paper's method in specification files, records experiment executions and provenance, keeps generated outputs separate from paper-provided assets, records comparison evidence between each generated output and the corresponding paper claim, and records coverage of the matched evidence in the final replication report before completion.
The replication workspace is the record of the run.
Validation checks determine whether recorded evidence qualifies as reproduced.
Because paper claims can be numeric, distributional, structural, or visual, the accepted evidence is claim-specific rather than exact numerical equality.
Completion is therefore a workspace state relative to the recorded target set and the validation checks applied to it.

We evaluate Paper-replication on twelve independent runs across four scientific machine learning papers: physics-informed information field theory (PIFT)~\citep{alberts2023physics}, data-driven solutions with physics-informed neural networks (PINN-I)~\citep{raissi2017physics}, data-driven discovery with physics-informed neural networks (PINN-II)~\citep{Raissi2017PhysicsID}, and sparse identification of nonlinear dynamical systems (SINDy)~\citep{brunton2016discovering}.
Across these runs, the workspaces record learned fields, solution errors, discovered coefficients, posterior structure, sparse supports, trajectories, and figures shaped by stochastic sampling, neural-network optimization, or simulated dynamics.
All twelve runs reach a completed workspace state under the completion gate.
Across repeated runs, the stored records also show variation in target decomposition, paper-anchored numeric fidelity, elapsed replication time, correction work, and recorded acceptance-rule type.

Our contributions are threefold.
First, we define paper replication as a target-level evidence task for a coding agent.
Each target must link a generated output to the paper claim, the agent's reconstruction of the paper's method, a successful run, provenance, comparison evidence, and report coverage before it can be marked as matched.
Second, we implement this task as Paper-replication, a coding-agent skill with a persistent replication workspace and validation checks.
The workspace records the target set, the agent's reconstruction of the paper's method and assumptions, run records, provenance records, comparison evidence, report coverage, and completion state, while the checks determine whether the recorded evidence can count as reproduced.
Third, we define and apply a repeated-run case-study analysis for evaluating Paper-replication.
The analysis measures paper-level completion, target coverage, paper-anchored numeric fidelity, elapsed replication time, correction work, and judgment variation across independent agent runs.

\section{Methods}\label{sec:method}
We use the term \emph{paper} to refer to scientific papers whose reproducible claims are computational: they focus on mathematical techniques, their implementation, and the results of computations.
This scope includes papers that use previously collected experimental data as inputs to those computations.
It excludes reproducing the physical process by which such data were collected.

We define paper replication as the task of reconstructing a paper's method from its text, regenerating the results that support its claims, and recording evidence for each claim.
We call the workflow that enforces this setting Paper-replication and implement it as a coding-agent skill.
The agent receives a prompt that instructs it to use Paper-replication and specifies the paper's \LaTeX{} source, datasets provided by or referenced in the paper, the compute environment, and policy choices such as whether author code is allowed.
In our setting, author code is not allowed.
The agent must therefore infer the method, implement it, run computational experiments, and prepare a report comparing its results with the paper's claims.

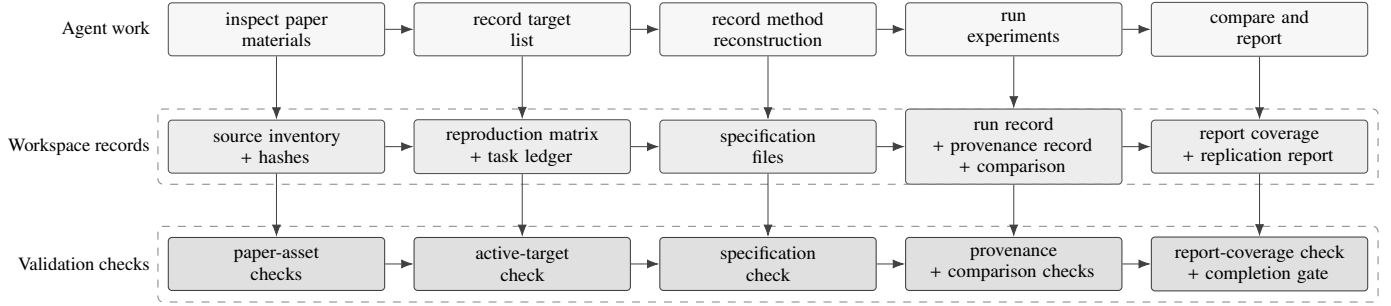
\begin{figure*}[t]
\centering
\begin{tikzpicture}[
  font=\scriptsize,
  >=Latex,
  x=1cm,
  y=1cm,
  action/.style={draw=black!68, rounded corners=1.4pt, fill=black!3, align=center, inner xsep=3pt, inner ysep=3pt, minimum height=7mm, text width=2.65cm},
  record/.style={action, fill=black!7},
  check/.style={action, fill=black!12, draw=black!75},
  lane/.style={draw=black!45, rounded corners=2pt, dashed},
  arr/.style={->, line width=.36pt, draw=black!75}
]
\node[anchor=east, align=right] at (-1.55,0) {Agent work};
\node[anchor=east, align=right] at (-1.55,-1.55) {Workspace records};
\node[anchor=east, align=right] at (-1.55,-3.10) {Validation checks};

\node[action] (a1) at (0,0) {inspect paper\\materials};
\node[action] (a2) at (3.25,0) {record target\\list};
\node[action] (a3) at (6.50,0) {record method\\reconstruction};
\node[action] (a4) at (9.75,0) {run\\experiments};
\node[action] (a5) at (13.00,0) {compare and\\report};

\node[record] (r1) at (0,-1.55) {source inventory\\+ hashes};
\node[record] (r2) at (3.25,-1.55) {reproduction matrix\\+ task ledger};
\node[record] (r3) at (6.50,-1.55) {specification\\files};
\node[record] (r4) at (9.75,-1.55) {run record\\+ provenance record\\+ comparison};
\node[record] (r5) at (13.00,-1.55) {report coverage\\+ replication report};

\node[check] (c1) at (0,-3.10) {paper-asset\\checks};
\node[check] (c2) at (3.25,-3.10) {active-target\\check};
\node[check] (c3) at (6.50,-3.10) {specification\\check};
\node[check] (c4) at (9.75,-3.10) {provenance\\+ comparison checks};
\node[check] (c5) at (13.00,-3.10) {report-coverage check\\+ completion gate};

\draw[lane] ($(r1.north west)+(-.13,.14)$) rectangle ($(r5.south east)+(.13,-.14)$);
\draw[lane] ($(c1.north west)+(-.13,.14)$) rectangle ($(c5.south east)+(.13,-.14)$);

\foreach \a/\b in {a1/a2,a2/a3,a3/a4,a4/a5,r1/r2,r2/r3,r3/r4,r4/r5,c1/c2,c2/c3,c3/c4,c4/c5}{\draw[arr] (\a) -- (\b);}
\foreach \a/\b in {a1/r1,a2/r2,a3/r3,a4/r4,a5/r5,r1/c1,r2/c2,r3/c3,r4/c4,r5/c5}{\draw[arr] (\a) -- (\b);}
\end{tikzpicture}
\caption{The Paper-replication workflow centers on two mechanisms: a persistent workspace that records the agent's state, and validation checks that decide whether recorded evidence can count as reproduced. The top row shows the work performed by the agent, the middle row shows the workspace records produced by that work, and the bottom row shows the validation checks applied to those records. The agent first inspects the paper materials and records the paper source, assets, and hashes. It then records the target list in the reproduction matrix, records its reconstruction of the paper's method in specification files, records experiment executions and provenance, and compares generated outputs with the paper's claims. The workspace records preserve these steps as the source inventory, reproduction matrix, task ledger, specification files, run record, provenance record, comparison evidence, report coverage, and replication report. The checks use the workspace records to test each step before completion: paper assets must remain separate from generated outputs, the task ledger and reproduction matrix must agree on the active target, specification files must record the agent's reconstruction of the paper's method, matched targets must have run-generated provenance and comparison evidence under the recorded acceptance rule, and matched evidence must appear in the report. The completion gate passes only when every recorded target has accepted evidence, no target remains active, and the final replication report PDF exists.}
\label{fig:paper_replication_workflow}
\end{figure*}

A prompt can state the Paper-replication goal, but a prompt alone does not give the agent a way to track progress or verify its own work.
With direct prompting, the agent often stops after reproducing only part of the paper, loses track of which claim to address next, or treats its own description of progress as evidence.
The agent can also count copied figures from the paper's \LaTeX{} source, figures produced to match the paper without implementing the method, or results from substitute methods as successful replications.
Paper-replication addresses these failure modes through harness engineering: it changes the environment in which the agent works by adding persistent workspace records and validation checks rather than relying on prompt instructions alone~\citep{lopopolo2026harness}.
First, it stores the agent's state in files under the replication workspace.
Second, it runs validation checks outside the agent before the agent can mark a claim as reproduced.
These files and checks record which claim the agent is working on, what evidence it has produced, and which checks must pass before the claim counts as reproduced.
To enable those checks, Paper-replication decomposes the paper-level replication task into explicit targets and evidence records for each target.
Figure~\ref{fig:paper_replication_workflow} summarizes the Paper-replication workflow.

\subsection{Replication targets and evidence}\label{sec:replication_problem}
Paper-replication makes the agent define the results and claims it must reproduce.
Let \(\mathcal{P}\) denote the paper materials available to the agent: the \LaTeX{} source tree, referenced figure and table assets, bibliography, appendices, and any datasets provided by or referenced in the paper.
The agent inspects \(\mathcal{P}\) and records each computational claim that requires reproduction as an element of a finite target set \(\mathcal{T}=\{t_j\}_{j=1}^{J}\).
A target is a result or claim selected for reproduction; it may correspond to a figure, table, reported scalar, learned field, empirical or posterior distribution, trajectory, discovered equation, or structural statement.
For each target, the agent records where the claim appears in the paper, which parts of the \LaTeX{} source or paper assets support its reconstruction, which data and method component it must use, what output the run must produce, how that output will be judged, what status the target currently has, and where the result must appear in the replication report.
These fields make each claim traceable from the paper source to the required method component, generated output, acceptance rule, and final report.
This traceability prevents an output from satisfying a target unless the agent links it to the paper's method and to the required comparison.
The reproduction matrix stores one record for each target, so progress and evidence are defined at the target level.

After the agent records what a target requires, Paper-replication ties the candidate result to the records needed to evaluate it.
For each target \(t_j\), let
\begin{equation}
\widehat{y}_j = F_j\!\left(D_j;\theta_j,\omega_j\right)
\label{eq:target_generation}
\end{equation}
denote the candidate result produced by the agent's reconstructed implementation \(F_j\) from input data \(D_j\), recorded configuration \(\theta_j\), and random seed or stochastic state \(\omega_j\).
Let \(y_j^{\star}\) denote the corresponding quantity or property reported in the paper.
It may be a value in the text, a table entry, a distributional property, or a structural feature of a figure rather than a stored numeric array.
For each target, Paper-replication requires the agent to record an evidence bundle
\begin{equation}
E_j = \left(\widehat{y}_j, R_j, P_j, C_j, G_j\right),
\label{eq:evidence_bundle}
\end{equation}
where \(R_j\) is the execution record for the run that produces the output, \(P_j\) links that output to the code, configuration, seed, and paper passages or equations used to justify the implementation, \(C_j\) compares \(\widehat{y}_j\) with \(y_j^{\star}\) under the target's acceptance rule, meaning the criterion the agent assigns to that target based on the claim's scientific content, and \(G_j\) records inclusion of the result in the replication report.
Paper-replication allows a target to be marked as matched only when every required part of its evidence bundle exists and the external checks accept it.
An output artifact alone, such as a generated figure or table, does not count as evidence.

The target and evidence formulation depend on the paper materials and the computing environment available to the agent.
It requires the agent to be able to identify and inspect \(\mathcal{P}\), and it requires each dataset to be included with the paper materials or described well enough for the agent to generate it when the experiment uses synthetic data.
It also requires a computing environment capable of running the experiments needed to reproduce the paper's claims.
The formulation does not assume that the paper states every implementation detail, that missing choices such as seeds, tolerances, initialization, or plotting conventions are uniquely determined, or that exact numerical or visual agreement is attainable.
When the paper omits details on preprocessing, architecture, optimization, sampling, or plotting, Paper-replication prompts the agent to treat them as hypotheses to test and record.
A target can be matched only when those hypotheses lead to a runnable implementation and a recorded comparison between \(\widehat{y}_j\) and \(y_j^{\star}\) that satisfies the target's acceptance rule.
When required data, compute resources, or method details remain unavailable, Paper-replication does not allow the target to be marked as matched.
The target remains unmatched, and the agent records the reason in the workspace files and replication report.

\subsection{Persistent workspace}\label{sec:persistent_workflow}
Paper-replication implements this target and evidence contract through a workspace for each paper.
A manifest file records the paper source, a hash used to detect source changes, run rules such as author-code access, the available compute environment, and how to rerun the workflow.
These entries let the agent resume after an interruption without changing the paper source or run conditions.

Before the target list is finalized, Paper-replication makes the agent inspect \(\mathcal{P}\) and build a source-based record of it.
The agent starts from the main \LaTeX{} file and records an inventory of included \LaTeX{} files, bibliography entries, appendices, figure assets, and data references.
The agent also records relevant unreferenced \LaTeX{} files whose names suggest supplementary methods or appendices, because they can contain details outside the main \LaTeX{} file.
This inventory helps the agent use the paper source as a guide for creating the target list, reconstructing the method, and later checking whether any paper claim was omitted.
When the environment permits, the agent compiles the paper into a PDF and converts the pages into images.
The rendered pages help the agent find claims that are visible in the paper but not easy to identify from the source files alone, including figure trends, subfigure labels, table entries, equations, and results embedded in image files.
The agent uses these pages to check that such claims appear in the target list and to assign comparison criteria for them.
During source inspection, Paper-replication records hashes of the paper source, referenced assets, and rendered pages.
These hashes allow later checks to detect whether generated output was copied from paper-provided material.
Hash checks do not rule out every transformed copy or adversarial reuse of paper-provided material by the agent.
In these runs, we did not observe such reuse.

After source inspection, the agent writes the target list to the reproduction matrix, which stores each target and its current status.
The agent updates this matrix as each target's status changes.
A task ledger records the one target currently being worked on, open questions, checks that must pass, and completed work.
Together, the matrix and ledger keep the agent focused on one result at a time and make unfinished work visible.

Once the target list is in the reproduction matrix, Paper-replication requires a recorded reconstruction of the paper's method before any target can be marked as matched.
Specification files record the target definitions, the agent's restatement of the paper's equations and algorithms, the implementation plan, assumptions and unresolved details, and notes about the paper's figures and tables.
These files make the agent's reconstruction choices inspectable instead of leaving them in the conversation.
The equation and algorithm notes tie the implementation to the paper's stated method, helping the agent distinguish a paper-faithful implementation from a substitute method.
The figure and table notes help the agent identify claims that appear only in visual or tabular form, such as trends, regimes, error patterns, or structural features.
When details are missing, the agent records each missing detail as a question, the assumption it used to answer that question, the check or experiment used to test the assumption, and the evidence that supports the final choice.
This turns guesses into recorded hypotheses.

With the target list and method specification in place, the agent then works through the target set, one target at a time.
A target starts as planned, becomes active while the agent implements and tests it, and becomes matched, recorded as \(\mathrm{MATCHED}\), only after its evidence bundle passes the external checks.
Targets that the workflow does not reproduce remain unmatched and do not count as matches.
The task ledger and reproduction matrix must name the same active target, and a validation check rejects multiple active targets or the absence of an active target while unresolved work remains.
This rule limits interference among partial experiments and provides the agent with a stopping condition: it must close the evidence contract for the current target before moving to the next.

To connect each result to the computation that produced it, every substantive execution passes through a run recorder.
The recorder stores the command used for the run, its working directory, start and finish times, whether the run succeeded, messages and errors produced during the run, the outputs it was expected to produce, and hashes of the produced files.
Failed runs and runs later replaced by corrected outputs remain in the record rather than being overwritten, so the workspace preserves the trial-and-correction path.
When a run produces a candidate output, the agent registers that output against the target and the successful run identifier.
The provenance record links the output to the implementation file, configuration file, seed, method components, and paper passages or equations used to justify the implementation.
It also stores hashes of the output, implementation file, and configuration file so later edits cannot silently change the output or implementation associated with the claim.
Code and configuration are stored separately from generated outputs, and paper-provided assets are stored separately from both.
This separation helps prevent the agent from treating a paper-provided figure or table as a generated result.
Together, the run record and provenance record allow a later validator to determine which executable choices produced the claimed result and whether those choices changed after the claim was recorded.

Because these records are stored in the workspace, a status check can recover the active target, unmatched targets, validation errors, and next action.
Paper-replication therefore treats the workspace, not the chat transcript, as the record of the run.

\subsection{Validation and completion}\label{sec:external_validation}
The workspace records progress, but recorded progress alone does not show that an output supports a paper claim.
Paper-replication therefore requires a validation check before the agent can mark a target as matched.
The check uses the target's acceptance rule, which the agent infers from the claim's scientific content.
The agent records the acceptance rule in the target record before judging the output, using any tolerance or accuracy class stated in the paper.
When the paper does not state such a criterion, the agent records the convention it uses and the reason for using it.
This prevents the agent from treating a plausible-looking figure, table, or scalar as evidence unless the recorded comparison supports the target, and it prevents the agent from changing the judgment rule after seeing a failed comparison.
The form of the acceptance rule depends on the kind of target.

For a numeric target, the acceptance rule records the discrepancy metric, its units, and the tolerance used to judge the result.
The candidate result is accepted only when the measured discrepancy between \(\widehat{y}_j\) and \(y_j^{\star}\) is no larger than that tolerance.
Numeric targets are claims that can be verified with numbers, such as a reported error, an estimated coefficient, a final loss, or a convergence rate.

Targets that cannot be judged by a single numerical comparison use acceptance rules that match the kind of claim being reproduced.
For a distributional target, the acceptance rule compares the properties of the distribution reproduced by the agent with those reported or shown in the paper, rather than requiring identical samples.
This rule applies to posterior samples, empirical distributions, uncertainty bands, and stochastic sampling behavior.
Depending on the claim, the agent records summary statistics such as the mean or variance, quantiles, interval coverage, support, number of modes, or a distance between distributions.
If that comparison requires a threshold for acceptance, the agent records the threshold in the target record before judging the output.
For a structural target, the acceptance rule records the pattern or relationship that the reproduced result must show, such as sparsity, ordering, stability, a regime transition, phase-portrait geometry, attractor structure, or selective identifiability.
For structural targets, the recorded comparison indicates which structural properties agree with the paper and which do not, and uses numerical summaries when available.
For a visual target, the output's appearance is part of the claim.
The evidence therefore records the reference output, the candidate output, and the visual comparison used to judge the match.
Paper-replication uses visual matching only when the claim requires it, because stochastic trajectories, convergence curves, and sampling-based figures can support the same claim without pixelwise agreement.

After the target-specific comparison is recorded, the external checks enforce the target-type requirements.
They reject visual-only evidence for numeric and distributional targets, missing visual comparisons for visual targets, and structural comparisons that do not explain how the reproduced structure agrees or disagrees with the paper.
The checks also address copied paper material and method substitution, because an output can appear correct without being derived from a reconstruction of the paper's method.
Paper-provided figures, rendered pages, and paper-source files remain separate from generated outputs.
A matched output must come from the replication workspace and must not match the paper source or paper-asset records.
Hash checks compare generated outputs with indexed paper assets and rendered pages and detect direct reuse of paper-provided material.
Because hash checks do not detect every transformed copy, Paper-replication also requires the recorded comparison for the target and the run provenance for the generated output.

For method-replication targets, the provenance record must show how the run implements the paper's method before the target can be matched.
The agent records the implemented method components, the paper passages or equations that justify those components, and the code, configuration, and seed used for the run.
This makes the replication claim depend on a trace from the paper's method to the executable implementation, not only on similarity between the generated output and the reported result.
If the agent cannot justify a method component from the paper materials or from a recorded assumption, the run cannot serve as evidence for the replication target.
The final replication report may include the run only as a deviation or comparison, not as evidence that the target was matched.
The external checks reject missing method provenance and records that support a claim only by showing that the output resembles the paper.
These checks reduce the risk that the agent reuses a figure or table provided in the paper or substitutes another method, but they do not prove that the agent's implementation is uniquely determined by the paper.
Paper-replication therefore treats a passed check as recorded evidence for the target, not as proof that no other implementation choices could also be consistent with the paper.

After target evidence passes validation, a report-coverage check connects that evidence to the paper-level replication claim.
For each matched target, the agent records where the generated output or value appears in the final replication report.
The report also records deviations, unmatched targets, and hardware or runtime conditions.
If a target remains unmatched, the agent records the reason in the workspace files and the report.
This prevents the agent from converting missing evidence into a successful paper-level claim.

These report-coverage requirements are part of the final completion gate.
Let \(V_{\mathrm{spec}}\), \(V_{\mathrm{progress}}\), and \(V_{\mathrm{report}}\) denote the validation checks for specification completeness, consistency of target state and evidence, and report coverage.
In addition to these three checks, completion requires that every target be matched, that no target remain active, and that the final report PDF exist.
Paper-replication allows the workspace to be marked complete only when
\begin{equation}
\begin{aligned}
V_{\mathrm{complete}} ={}& V_{\mathrm{spec}} \wedge V_{\mathrm{progress}} \wedge V_{\mathrm{report}} \\
& \wedge \left(\bigwedge_{j=1}^{J} [s_j=\mathrm{MATCHED}]\right) \wedge [a=\varnothing] \\
& \wedge [\text{report PDF exists}].
\end{aligned}
\label{eq:completion_gate}
\end{equation}
Here \(J\) is the number of recorded targets, \(s_j\) is the recorded status of target \(j\), \(a\) is the active-target field, \(\varnothing\) denotes no active target, and bracketed expressions denote Boolean indicators.
This gate makes completion a workspace state rather than a statement in the agent's final message.
A run may end with a documented non-reproduction, but Paper-replication cannot mark the workspace as complete while any target lacks accepted evidence.
Completion remains relative to the target set recorded in the reproduction matrix.
During final review, the agent compares the reproduction matrix with the source inventory and rendered pages, adds any omitted computational claim as a target or records why it is out of scope, and then reruns the completion checks.
The workflow does not guarantee that every paper can be replicated from the available materials, and an accepted target may still omit aspects of the paper's claim that were not included in its acceptance rule.
Within these limits, replication is complete only when every recorded target has accepted evidence, appears in the final replication report, and the report exists as a rendered PDF.

\subsection{Skill implementation}\label{sec:skill_implementation}
The implemented skill has two layers: persistent agent instructions and replication workspace utilities.
The instruction layer is defined in \texttt{SKILL.md}, while separate Codex and Claude Code prompt files adapt those instructions to each agent interface.
These instructions tell the agent to initialize or reopen the replication workspace before substantive work, start from the paper's \LaTeX{} source, use the available compute environment and the specified author-code rule, keep exactly one active target, and treat the workspace files rather than the chat transcript as the record of the run.
Files under \path{references/} define the workspace contract, author-code rule, available compute environment, and acceptance-rule types used by the validation checks.
These files do not contain paper-specific scientific instructions.
Instead, they state the workflow rules in a form the agent can reopen after an interruption or after losing earlier chat context.
The instruction files define the rules the agent must follow.
The workspace utilities create and check the files that make those rules enforceable.

The workspace utilities first create the files that make the workflow recoverable.
They are implemented in \path{scripts/paper_replication.py}.
Initialization and synchronization commands create the workspace directories and write the manifest, reproduction matrix, task ledger, specification files, and replication report template.
The manifest records the paper title, paper source, a hash used to detect source changes, run rules such as author-code access, the available compute environment, and how to rerun the workflow.
The reproduction matrix stores, for each target, where the claim appears in the paper, which source files or assets support it, which run and configuration produce the output, how the output will be judged, the target's current status, and where the result appears in the replication report.
The task ledger stores the active target, open questions, checks that must pass, and completed work.
Because these files live in the workspace, the status check can recover the active target, unmatched targets, validation errors, and next action without relying on the chat transcript.

Once the workspace exists, the agent uses the utilities to record the paper source and keep paper-provided material separate from generated outputs.
During source inspection, the agent starts from the main TeX file and records included TeX files, bibliography entries, appendix files, figure assets, and data references in \path{spec/paper_inventory.json}.
The workspace utilities also index paper-provided figures and source files under \path{artifacts/paper_figures/} and record their hashes.
When the environment permits, the agent uses these utilities to compile the paper and render pages so visible claims can be checked against the target list.
Generated outputs are stored separately under \path{artifacts/figures/} and \path{artifacts/tables/}.
This directory-and-hash separation lets later checks reject a candidate output that reuses a paper-provided asset, rendered page, or file from the paper's source tree.

After the paper source is recorded, the same utilities record the computations that generate candidate evidence.
When the agent runs an experiment, it invokes the run recorder with the command for that run.
The recorder stores the command, working directory, start and finish times, whether the run succeeded, messages and errors produced during the run, expected outputs, and hashes of the produced files in \path{artifacts/runs/}.
When a run produces a candidate output, the agent registers that output against the target and the successful run identifier.
The workspace utilities then write a provenance record under \path{artifacts/provenance/}.
This record identifies the method components the agent implemented and links the output to the implementation file, configuration file, seed, successful run, and paper passages or equations used to justify the implementation.
It also stores hashes of the output, implementation file, and configuration file so later edits cannot silently change the output or implementation associated with the claim.
The utilities do not implement the source paper's method for the agent.
Instead, they check that the provenance record exists and is consistent with the registered target output.
For a target to be matched, this record must show how the implementation follows the paper's method, not only how the agent generated an output resembling the paper's reported result.
These records reduce the risk that output similarity alone counts as evidence for a matched target.

The recorded workspace state then becomes the input to the external checks that the completion gate uses.
The specification check verifies that the manifest, task ledger, paper inventory, source hashes, specification files, and reproduction matrix exist and contain the required records.
The progress check verifies consistency between the task ledger and the reproduction matrix, confirms that matched targets have run-generated provenance, rejects copied paper material based on path and hash, and enforces the comparison evidence required by each target's acceptance rule.
The report-coverage check verifies that matched figure and table targets appear in \path{report/main.tex} at their recorded report locations.
The completion check runs these checks and then requires that every target be matched, that no target remain active, and that \path{report/main.pdf} exist.
The agent invokes these checks, but completion is a workspace state rather than a statement in the agent's final message.
Together, these checks preserve the two workflow principles used throughout Paper-replication: the run state persists in workspace files, and evidence counts only when workspace records and validation checks outside the agent accept it.

\subsection{Case study analysis}\label{sec:method_analysis}
We apply Paper-replication to four scientific machine learning papers: PIFT, PINN-I, PINN-II, and SINDy.
We choose these papers because together they require different kinds of target evidence, including learned fields, solution errors, discovered coefficients, posterior structure, sparse supports, trajectories, and figures shaped by stochastic sampling, neural-network optimization, or simulated dynamics.
For each paper, we run three independent agent replications with the same paper materials, initial and follow-up prompt templates, available tools, compute environment, and policy that excludes author code.
The workspaces do not share code, generated outputs, or persistent state, so each run reconstructs the paper independently.
This lets us measure variation across independent agent runs using the same Paper-replication workflow.

Each run begins with an initial prompt that specifies the paper source, replication workspace, available compute environment, and author-code rule, and instructs the agent to use Paper-replication until the completion gate passes.
Because a long-running agent can still stop before the workspace is complete, each run also has a fixed queue of ten follow-up prompts.
A follow-up prompt instructs the agent to reopen the same replication workspace, run the status check, and continue until all recorded targets are matched and the completion gate passes.
The queue allows the agent to resume from the same workspace if it stops before completion.
If the gate has already passed, the agent reports the completed workspace state.
Thus, the follow-up prompts support continuation, while completion remains defined by the workspace state.

The resulting workspaces provide the records used for the case-study analysis.
Let \(p=1,\ldots,N\) index papers and \(r=1,\ldots,R\) index runs, with \(N=4\) and \(R=3\).
For each workspace, we use the final reproduction matrix, target statuses, acceptance rules, comparison records, run records, execution outcomes, elapsed replication time, and report coverage.
We call a run paper-level complete when the completion gate in Eq.~\eqref{eq:completion_gate} passes.
If a run does not pass this gate, we retain it as incomplete, along with its validation errors and the reasons for unmatched targets.
Beyond completion, we analyze target coverage, paper-anchored numeric fidelity, effort, and variation in judgment.

Repeated runs may organize the same paper differently.
They may assign different target identifiers or split the same paper result into different numbers of targets.
For cross-run comparisons, we therefore group targets by the paper they support.
The grouping uses the recorded paper location and scientific content, such as a figure, table, equation, caption statement, reported scalar, or paragraph-level claim.
This lets us compare repeated runs even when their reproduction matrices are not organized in the same way.

We use \(c\) to index the aligned paper claims produced by this grouping.
The models below use \(p\) for papers and \(r\) for runs within a paper.
A subscript \(pr\) denotes a run-level quantity for paper \(p\), \(pc\) denotes a claim-level quantity for aligned claim \(c\) in paper \(p\), and \(pcr\) denotes a run-level quantity for aligned claim \(c\) in run \(r\) of paper \(p\).
Symbols without these indices denote corpus-level parameters shared across papers.

We first analyze variation in how the agent decomposes each paper into targets.
Target coverage is the number of targets \(Q_{pr}\) in the final reproduction matrix for run \(r\) of paper \(p\).
It measures how finely the agent decomposes the paper's computational claims into separate targets.
It does not measure correctness because every listed target must still satisfy the evidence contract and pass the external checks.
We model target coverage with a Gamma-Poisson likelihood, parameterized by concentration and rate:
\begin{equation}
\begin{gathered}
Q_{pr}\mid \lambda_p,\phi \sim \mathrm{GammaPoisson}(\phi,\phi/\lambda_p),\\
\log \lambda_p = \mu_Q + a_p,\\
a_p\mid \sigma_Q \sim \mathrm{Normal}(0,\sigma_Q^2), \qquad
\mu_Q \sim \mathrm{Normal}(2.5,1^2),\\
\sigma_Q \sim \mathrm{HalfNormal}(1), \qquad
\phi \sim \mathrm{HalfNormal}(10).
\end{gathered}
\label{eq:coverage_model}
\end{equation}
Here \(Q_{pr}\) is the observed target count for run \(r\) of paper \(p\), while \(\lambda_p\) is the expected target count for paper \(p\) and \(a_p\) is that paper's deviation from the corpus mean.
Because target coverage is a run-level count, the model has no claim-level term.
The corpus-level parameters are the overall log target count \(\mu_Q\), the between-paper scale \(\sigma_Q\), and the shared overdispersion parameter \(\phi\).
We also report the decomposition ratio, defined for each paper as the largest final target count divided by the smallest final target count across its three runs.
This ratio gives a direct summary of how much the agent's target decomposition varies for the same paper.

We next evaluate numerical fidelity for aligned claims with scalar paper-reported quantities.
For scalar claims, we define standardized numeric anchors.
A standardized numeric anchor is an aligned paper claim with a reproducible scalar quantity, such as a reported relative \(L_2\) field error or the percentage error of an identified coefficient.
For paper \(p\), aligned claim \(c\), and run \(r\), let \(d_{pcr}\geq 0\) denote the discrepancy between the quantity reproduced by the agent and the value or accuracy class reported in the paper.

Each anchor receives a fixed paper-anchored threshold \(\tau_{pc}>0\).
We choose \(\tau_{pc}\) from the accuracy scale reported for that paper claim and fix it before comparing repeated runs.
We express \(d_{pcr}\) in the same units as \(\tau_{pc}\).
This threshold does not depend on the acceptance rule, metric name, or tolerance recorded by the agent in an individual workspace.
For PINN-I and SINDy, the thresholds are taken directly from the reported error scales in the source papers.
For PINN-II coefficient identification, we use a ten-percent coefficient-error threshold.
The paper reports clean-data coefficient errors below five percent and evaluates robustness to up to ten percent data noise, so we use ten percent as an analysis convention for the reported coefficient-accuracy class.
This threshold is an analysis convention based on the paper's reported robustness scale, not a claim that data noise and coefficient error are equivalent.
Table~\ref{tab:case_study_thresholds} summarizes the analysis rules used for the case-study claims.

\begin{table}[t]
\centering
\footnotesize
\setlength{\tabcolsep}{3pt}
\renewcommand{\arraystretch}{1.08}
\begin{tabularx}{\columnwidth}{@{}l
  >{\raggedright\arraybackslash}X
  >{\raggedright\arraybackslash}X
  c@{}}
\toprule
Paper & Claim family & \makecell[l]{Paper-anchored\\analysis rule} & \makecell{Analysis\\type} \\
\midrule
PINN-I & solution relative \(L_2\) error & \(d_{pcr}\le 10^{-2}\) & scalar \\
PINN-II & coefficient percentage error & \(d_{pcr}\le 10\%\) & scalar \\
SINDy & Lorenz coefficient relative error & \(d_{pcr}\le 10^{-3}\) & scalar \\
\midrule
SINDy & sparse-support recovery and trajectory geometry & exact support or structural agreement, depending on the target & structural \\
PIFT & posterior collapse, bimodality, and selective identifiability & distributional or structural evidence, depending on the target & non-scalar \\
\bottomrule
\end{tabularx}
\caption{Analysis rules used in the case-study analysis. The scalar thresholds remain fixed across runs and do not depend on the tolerance recorded by the agent in an individual workspace. Claims without a paper-reported scalar are analyzed through the structural or distributional evidence required by their target acceptance rules.}
\label{tab:case_study_thresholds}
\end{table}

To compare heterogeneous scalar quantities on one scale, we define \(\widetilde{d}_{pcr}=\max(d_{pcr},\epsilon_{pc})\), where \(\epsilon_{pc}\) is a claim-specific resolution floor held fixed across runs.
We set \(\epsilon_{pc}\) to the smallest positive discrepancy observed in the stored results for that anchor.
This floor prevents an exact recorded equality from producing infinite headroom.
We then define
\begin{equation}
h_{pcr} = \log_{10}\!\left(\frac{\tau_{pc}}{\widetilde{d}_{pcr}}\right).
\label{eq:headroom}
\end{equation}
Positive headroom means that the reproduced discrepancy lies inside the fixed threshold, zero means that it lies on the threshold, and negative headroom means that it lies outside.
One unit of headroom represents a tenfold change in discrepancy relative to the threshold.

The scalar anchors vary across papers, aligned claims within a paper, and repeated runs.
To describe these sources of variation, we model headroom with paper-level, claim-level, and run-level terms:
\begin{equation}
\begin{gathered}
h_{pcr}\mid \mu,b_p,g_{pc},\sigma_{\varepsilon,p}
\sim \mathrm{Normal}(\mu+b_p+g_{pc},\sigma_{\varepsilon,p}^2),\\
b_p\mid\sigma_b \sim \mathrm{Normal}(0,\sigma_b^2), \qquad
g_{pc}\mid\sigma_g \sim \mathrm{Normal}(0,\sigma_g^2),\\
\mu \sim \mathrm{Normal}(0,3^2), \qquad
\sigma_b,\sigma_g,\sigma_{\varepsilon,p}
\stackrel{\mathrm{ind}}{\sim}\mathrm{HalfNormal}(1.5).
\end{gathered}
\label{eq:headroom_model}
\end{equation}
Here \(h_{pcr}\) is the observed headroom for aligned claim \(c\) in run \(r\) of paper \(p\).
The paper-level term \(b_p\) shifts the average headroom for paper \(p\), and \(\sigma_{\varepsilon,p}\) is the paper-specific residual scale across repeated runs.
The claim-level term \(g_{pc}\) captures variation among aligned claims within paper \(p\).
The corpus-level parameters are the mean headroom \(\mu\), the between-paper scale \(\sigma_b\), and the between-claim scale \(\sigma_g\).
Because \(h_{pcr}=0\) corresponds to the fixed threshold, the prior \(\mu\sim\mathrm{Normal}(0,3^2)\) does not assume that reproduced quantities usually fall inside or outside the threshold.
The prior scale allows discrepancies several orders of magnitude above or below the threshold.
From this model, we also report the posterior predictive probability that another run of a scalar anchor has positive headroom.

Not every paper claim admits a scalar discrepancy.
Some claims concern distributions, such as posterior shape or bimodality.
Other claims concern qualitative structure in the reproduced equation, field, or trajectory.
For these targets, we report the comparison evidence required by the target's acceptance rule and do not force the result into the scalar headroom model.
This keeps the analysis aligned with the kind of claim the source paper makes.

We measure effort for each replication run as elapsed replication time.
For each workspace, replication time starts when the initial replication prompt is given to the agent and ends when the agent first reports completion after the completion gate has passed.
Queued follow-up prompts sent after that point only confirm the completed workspace state, so they are not included in elapsed replication time.
Let \(H_{pr}>0\) denote this elapsed time in hours for run \(r\) of paper \(p\).
This quantity includes agent work, experiment execution, validation, report preparation, and waiting time before the first completion report.
Because elapsed times are positive and can vary by multiplicative factors, we model effort as \(\log_{10}H_{pr}\):
\begin{equation}
\begin{gathered}
\log_{10} H_{pr}\mid \mu_H,u_p,\sigma_H
\sim \mathrm{Normal}(\mu_H+u_p,\sigma_H^2),\\
u_p\mid\sigma_u \sim \mathrm{Normal}(0,\sigma_u^2),\\
\mu_H \sim \mathrm{Normal}(0,1.5^2), \qquad
\sigma_u,\sigma_H \stackrel{\mathrm{ind}}{\sim}\mathrm{HalfNormal}(1).
\end{gathered}
\label{eq:effort_model}
\end{equation}
Here \(\log_{10}H_{pr}\) is the observed log elapsed time.
The paper-level term \(u_p\) shifts the average effort for paper \(p\).
Because elapsed effort is defined for the workspace as a whole, the model has no claim-level term.
The corpus-level parameters are the overall mean log effort \(\mu_H\), the between-paper scale \(\sigma_u\), and the shared run-to-run residual scale \(\sigma_H\) after conditioning on paper.

Replication time gives one measure of effort.
As a second measure, we use the run records to count tracked executions and correction work.
A tracked execution is any recorded command used to generate, check, or prepare target outputs for the replication report.
We call an execution superseded when it fails or when no final target provenance record links its output to a retained result because later correction work replaces it.
Such corrections arise when the agent misses a method detail, resolves an underspecified step, rejects an initially plausible implementation choice, or replaces an unstable run.
We report the number of tracked executions and the number superseded as secondary effort diagnostics.
These counts show how much correction work occurs before the final evidence is accepted.

Judgment variation captures whether independent runs choose the same target-level acceptance rule for the same paper claim.
For each aligned claim, we compare the acceptance rule type recorded across runs.
We report the fraction of aligned claims for which every run of a paper records the same rule type.

We fit all Bayesian models in NumPyro with the No-U-Turn sampler, using four chains and two thousand retained posterior draws per chain.

\section{Results}\label{sec:results}
We evaluate Paper-replication using twelve independent replications drawn from four case-study papers, with three runs per paper.
Each run uses the same paper materials, prompt templates, compute environment, and no-author-code policy described in Section~\ref{sec:method_analysis}.
The coding agent is Codex with GPT-5.4 at the Extra High reasoning setting.
The agent runs from a MacBook Pro with an M4 Max chip and \(128~\mathrm{GB}\) of memory.
It also has access through the cluster-execution skill to Purdue Gautschi resources: CPU nodes with \(192\) cores and \(384~\mathrm{GB}\) of memory, and GPU nodes with eight H100 GPUs, \(112\) CPU cores, and about \(1~\mathrm{TB}\) of memory per node.
During each run, the agent chooses whether to keep work local or submit heavier jobs to the cluster.
In the observed runs, PIFT and SINDy complete on local compute, while some PINN-I and PINN-II targets use cluster jobs.
The resulting workspaces record local and cluster executions through the same run-record and provenance mechanisms.
This changes where work is executed, but it does not change the target-level evidence contract.

\begin{figure*}[t]
\centering
\includegraphics[width=\textwidth]{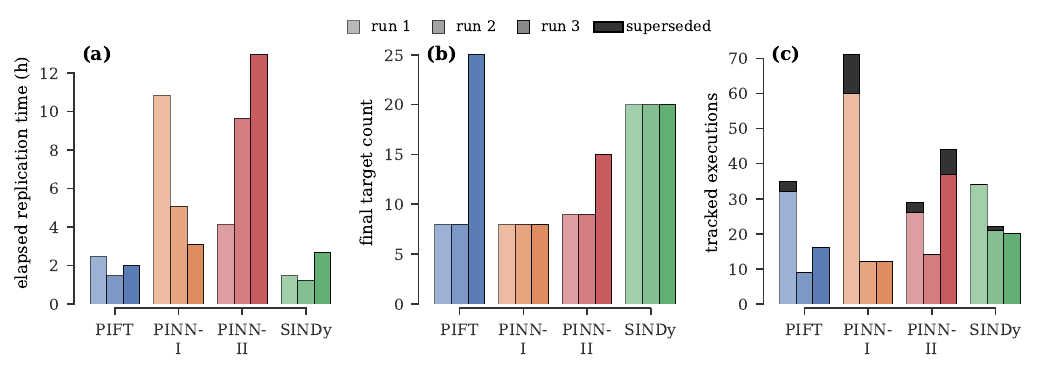}
\caption{Per-run workspace evidence for the twelve case-study runs. Panel (a) shows elapsed replication time. Panel (b) shows the final target count in the reproduction matrix. Panel (c) shows tracked executions, with dark segments indicating executions that were later superseded by correction work. Within each paper, lighter-to-darker shading denotes runs 1 to 3. Every run passes the completion gate, while target decomposition and recorded execution work vary across runs.}
\label{fig:dashboard}
\end{figure*}

We first evaluate paper-level completion under the completion gate in Eq.~\eqref{eq:completion_gate}.
Across the twelve final reproduction matrices, the agent records \(158\) targets.
Every recorded target reaches \(\mathrm{MATCHED}\), every matched target has report coverage, and every workspace passes the completion gate.
Paper-replication therefore reaches paper-level completion in all twelve runs, relative to the recorded target sets and the validation checks applied to them.
Table~\ref{tab:corpus} summarizes the completed workspaces, and Figure~\ref{fig:dashboard} shows the corresponding elapsed replication time, final target count, and tracked executions for each run.
Completion is stable across runs, while target decomposition, elapsed effort, and tracked executions vary by paper and by run.

\begin{table}[t]
\centering
\footnotesize
\setlength{\tabcolsep}{3pt}
\renewcommand{\arraystretch}{1.08}
\begin{tabular*}{\columnwidth}{@{\extracolsep{\fill}}l c c c c c@{}}
\toprule
Paper & Runs & Targets/run & Matched & \makecell{Elapsed\\time (h)} & \makecell{Superseded\\executions} \\
\midrule
PIFT & 3 & 8, 8, 25 & all & 2.2 [1.1, 4.4] & 3 \\
PINN-I & 3 & 8, 8, 8 & all & 5.0 [2.5, 9.9] & 11 \\
PINN-II & 3 & 9, 9, 15 & all & 6.9 [3.0, 13.4] & 10 \\
SINDy & 3 & 20, 20, 20 & all & 1.9 [1.0, 4.3] & 1 \\
\bottomrule
\end{tabular*}
\caption{Summary of the twelve case-study runs. Targets/run gives the number of targets in the final reproduction matrix for each of the three runs. Matched indicates that every recorded target reaches \(\mathrm{MATCHED}\) and the completion gate passes. Elapsed replication time gives the posterior median and \(95\%\) credible interval from the effort model. The last column counts tracked executions that later correction work replaces before final evidence is accepted.}
\label{tab:corpus}
\end{table}

\subsection{Target coverage and decomposition variation}\label{sec:results_coverage}
Target coverage is the number of targets in the final reproduction matrix for a run.
It measures how finely the agent decomposes a paper into recorded targets.
It does not, by itself, measure correctness because every recorded target must still satisfy its evidence contract and pass external checks.

\begin{figure}[t]
\centering
\includegraphics[width=\columnwidth]{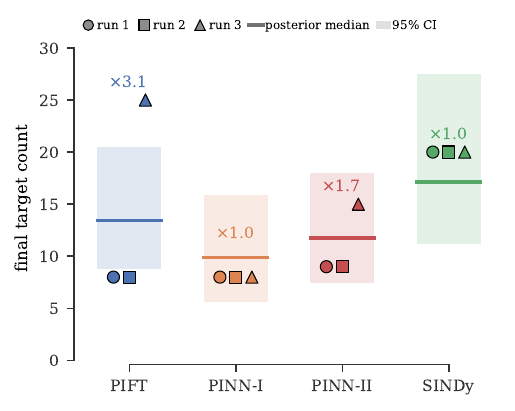}
\caption{Target coverage across runs. Points show the final target count recorded in each reproduction matrix. The horizontal bar and shaded interval show the posterior median and \(95\%\) credible interval from the Gamma-Poisson model. The annotation gives the decomposition ratio, defined as the largest final target count divided by the smallest final target count for the same paper. PIFT and PINN-II show run-dependent target decomposition, whereas PINN-I and SINDy do not.}
\label{fig:coverage}
\end{figure}

Two papers show no run-level variation in target coverage.
PINN-I records eight targets in all three runs, and SINDy records twenty targets in all three runs.
Two papers vary by run.
PIFT records eight, eight, and twenty-five targets, while PINN-II records nine, nine, and fifteen targets.
The resulting decomposition ratios are \(3.1\) for PIFT and \(1.7\) for PINN-II, compared with \(1.0\) for PINN-I and SINDy (Figure~\ref{fig:coverage}).
The Gamma-Poisson model in Eq.~\eqref{eq:coverage_model} reflects the larger spread for PIFT, whose expected target count has posterior median \(13.4\) with a \(95\%\) credible interval from \(8.8\) to \(20.5\).

The target-count differences come from decomposition choices rather than incomplete workspaces.
For cross-run comparison, we align targets by the paper claim they support rather than by target identifier.
Under that alignment, the extra PIFT targets occur when a single run records separate targets for different panels in a composite figure.
The extra PINN-II targets occur when one run records appendix inference claims as additional targets.
PINN-I and SINDy allow less variation of this kind because their reproducible claims are organized in the papers as named figures, tables, and examples.

\subsection{Paper-anchored numeric fidelity}\label{sec:results_fidelity}

\begin{figure*}[t]
\centering
\includegraphics[width=\textwidth]{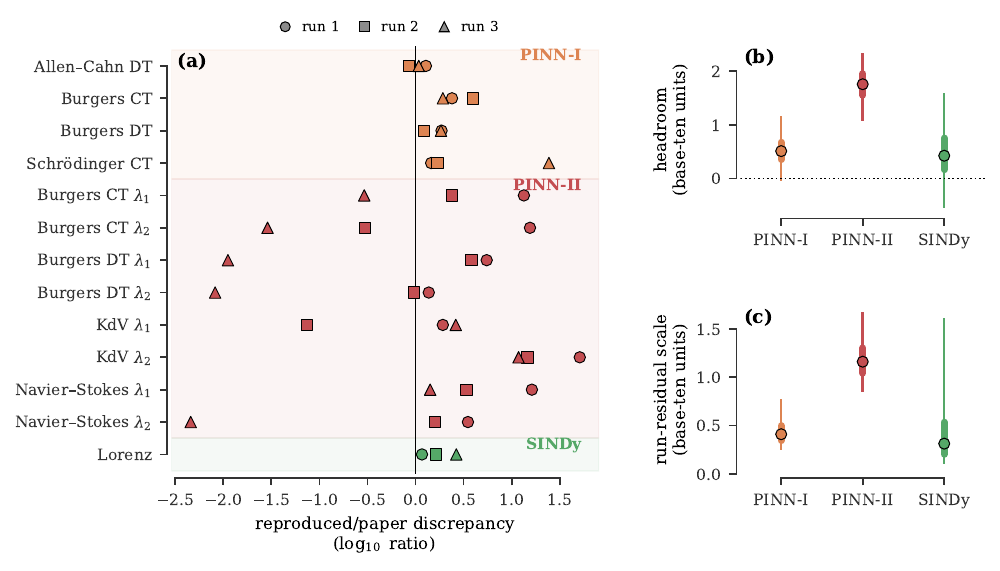}
\caption{Paper-anchored scalar fidelity for the thirteen standardized numeric anchors. Panel (a) shows the base-ten logarithm of the reproduced discrepancy divided by the paper-reported discrepancy for each anchor and run. Zero denotes equal discrepancy, positive values denote larger reproduced discrepancy, and negative values denote smaller reproduced discrepancy. This panel shows sensitivity to unspecified stochastic and implementation choices rather than the target-level acceptance rule. Panel (b) shows the posterior distribution of per-paper headroom, defined as the base-ten margin by which reproduced discrepancies fall inside the fixed paper-anchored threshold. Panel (c) shows the posterior distribution of the per-paper run-residual scale \(\sigma_{\varepsilon,p}\), which measures run-to-run movement in scalar fidelity. In panels (b) and (c), dots mark posterior medians, thick intervals mark interquartile ranges, and thin intervals mark \(95\%\) credible intervals. PIFT has no scalar anchor and is evaluated through distributional and structural evidence.}
\label{fig:fidelity}
\end{figure*}

We next evaluate scalar paper claims under paper-anchored thresholds.
This analysis is separate from workspace matching.
A workspace match uses the target-specific acceptance rule recorded in the reproduction matrix, whereas the paper-anchored analysis recomputes a standardized discrepancy for scalar anchors and compares it with a fixed threshold taken from the source paper's reported accuracy scale.

The scalar analysis contains thirteen standardized numeric anchors: four PINN-I solution errors, eight PINN-II coefficient errors, and one SINDy Lorenz coefficient error.
Across three runs per paper, these anchors yield thirty-nine anchor-run observations.
PIFT does not contribute a scalar anchor because its paper-level claims concern posterior collapse, bimodality, and selective identifiability rather than a paper-reported scalar error.
We therefore evaluate PIFT through the distributional and structural evidence recorded for its targets.

Of the thirty-nine scalar anchor-run observations, \(37\) fall inside the fixed paper-anchored threshold.
Two observations fall outside that threshold: Schr\"odinger run~3 has relative \(L_2\) error \(4.8\times 10^{-2}\) against a one-percent threshold, and Navier--Stokes \(\lambda_2\) run~1 has coefficient error \(16.4\%\) against a ten-percent threshold.
Both targets still count as \(\mathrm{MATCHED}\) in their workspaces because the workspace checks apply the acceptance rule recorded for that target.
They fall outside only under the separate paper-anchored scalar analysis.

The observed scalar discrepancies also show why exact equality to a printed number is not the relevant replication standard for these case studies.
The source papers often do not report the seeds, sampled training sets, optimizer states, or sampler states that would determine one stochastic realization.
Figure~\ref{fig:fidelity}a therefore reports the reproduced-to-paper discrepancy ratio on a base-ten scale, which makes sensitivity to those unspecified choices visible.
Some quantities move by orders of magnitude while remaining inside the source paper's reported accuracy class.
For example, the clean Burgers \(\lambda_2\) coefficient error takes values \(7.3\%\), \(0.14\%\), and \(0.014\%\) across the three runs, all below the ten-percent coefficient-error threshold used for PINN-II.
Several run~3 PINN-II coefficients are also more precise than the values printed in the paper.

The headroom model in Eq.~\eqref{eq:headroom_model} summarizes this variation on the fixed-threshold scale.
Average headroom remains positive for every paper with scalar anchors (Figure~\ref{fig:fidelity}b).
The posterior medians are \(0.51\) for PINN-I, \(1.75\) for PINN-II, and \(0.42\) for SINDy.
These values mean that reproduced discrepancies sit about \(3.2\times\), \(57\times\), and \(2.6\times\) inside their thresholds on average.
The corresponding posterior predictive probabilities that another anchor-run falls inside the paper-anchored threshold are \(0.79\), \(0.90\), and \(0.73\).
The same model also estimates the run-residual scale \(\sigma_{\varepsilon,p}\), which measures run-to-run movement in scalar fidelity on the base-ten discrepancy scale.
It is largest for PINN-II (Figure~\ref{fig:fidelity}c), with posterior median \(1.16\) and \(95\%\) credible interval \([0.86,1.66]\).
This corresponds to about a factor of \(14\) in discrepancy across reruns.
PINN-I has posterior median \(0.41\) with \(95\%\) credible interval \([0.26,0.76]\), while SINDy has posterior median \(0.31\) with \(95\%\) credible interval \([0.11,1.60]\).
The SINDy interval is wide because the scalar model contains one SINDy anchor.

For targets without scalar anchors, the comparison remains claim-specific rather than headroom-based.
For these targets, \(\mathrm{MATCHED}\) means that the recorded comparison satisfies the target-specific acceptance rule rather than the fixed paper-anchored threshold used in the scalar analysis.
For the PINN table targets, the reproduced tables preserve the reported accuracy patterns as data size, collocation points, network capacity, Runge--Kutta stages, and noise level change, even when individual cell values do not match every printed digit.
For SINDy, the agent recovers exact sparse support in every run for every system, and the trajectory figures preserve the attractor, bifurcation, and slow-manifold geometry required by the structural targets.
The cylinder-wake target shows why this distinction matters.
The paper materials describe direct numerical simulation followed by proper orthogonal decomposition, but they do not include the simulation trajectories, reduced-order trajectories, or the preprocessing conventions needed to verify digit-level coefficient equality.
The matched evidence therefore supports the quadratic wake structure and trajectory geometry rather than exact equality of coefficients.
For PIFT, the posterior standard deviation decreases monotonically as the inverse temperature increases across all three runs; the Allen--Cahn prior and posterior remain bimodal with exactly two modes; and the diffusion coefficient is sharply identified, whereas the reaction coefficient is not.

\subsection{Effort and correction work}\label{sec:results_effort}

\begin{figure*}[t]
\centering
\includegraphics[width=\textwidth]{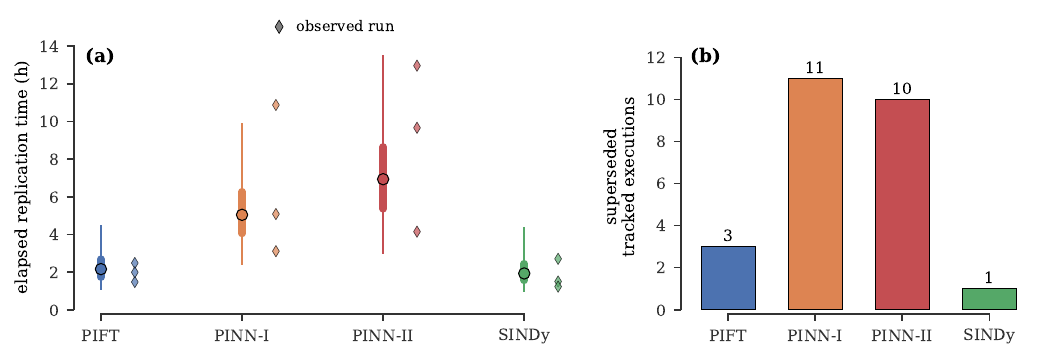}
\caption{Elapsed replication time and correction work across papers. Panel (a) shows the posterior distribution of per-paper elapsed replication time. The dot marks the posterior median, the thick interval marks the interquartile range, the thin interval marks the \(95\%\) credible interval, and diamonds mark observed elapsed replication times. Panel (b) counts superseded tracked executions summed over the three runs of each paper. Most superseded executions occur in the two PINN papers.}
\label{fig:effort}
\end{figure*}

Elapsed replication time is the workspace-level effort measure: it starts when the initial replication prompt is given to the agent and ends when the agent first reports completion after the completion gate has passed.
Paper-level completion is constant across the twelve runs, but elapsed effort is not.
The observed elapsed replication time ranges from \(1.2\) to \(13.0\) hours.
The effort model in Eq.~\eqref{eq:effort_model} estimates longer elapsed replication times for PINN-II and PINN-I than for PIFT and SINDy (Figure~\ref{fig:effort}a).
The posterior median elapsed replication time is \(6.9\) hours for PINN-II and \(5.0\) hours for PINN-I, compared with \(2.2\) hours for PIFT and \(1.9\) hours for SINDy.
For the four comparisons between a PINN paper and a non-PINN paper, the posterior probability that the PINN paper takes longer ranges from \(0.947\) to \(0.972\).
Even after conditioning on paper, repeated runs differ by about a factor of two in elapsed replication time.

The run records show one source of variation in this effort.
Across the corpus, the workspaces record twenty-five superseded tracked executions.
These are tracked executions that the agent initially uses when implementing or checking a target, but later replaces with correction work before final evidence is accepted.
Twenty-one of the twenty-five superseded executions occur in the two PINN papers: eleven in PINN-I and ten in PINN-II.
PIFT has three, and SINDy has one (Figure~\ref{fig:effort}b).
Within-paper effort also varies.
In PINN-I, run~1 records seventy-one tracked executions, eleven of which are superseded, whereas runs~2 and~3 each record twelve tracked executions with none superseded.
The longer PINN runs, therefore, reflect not only training and optimization cost, but also correction work before final evidence is accepted.

\subsection{Judgment variation in acceptance rule types}\label{sec:results_drift}
Judgment variation asks whether independent runs record the same target-level acceptance rule type for the same aligned paper claim.
For each aligned claim, we compare the recorded acceptance rule type across runs.
We count agreement only when every run of that paper records the same rule type.

\begin{figure}[t]
\centering
\includegraphics[width=\columnwidth]{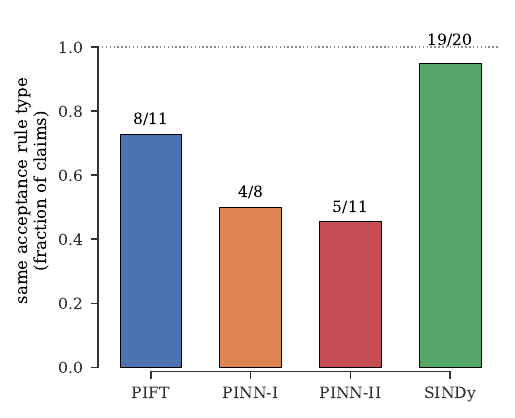}
\caption{Judgment variation across runs. Bars show the fraction of aligned paper claims for which every run of the same paper records the same target-level acceptance rule type. Labels give the number of aligned claims with agreement divided by the number compared for that paper.}
\label{fig:drift}
\end{figure}

Run-to-run agreement is highest for SINDy and lowest for PINN-II.
The same-rule fraction is \(19/20=0.95\) for SINDy, \(8/11=0.73\) for PIFT, \(4/8=0.50\) for PINN-I, and \(5/11=0.46\) for PINN-II (Figure~\ref{fig:drift}).
The PINN papers show greater variation in judgment because repeated runs sometimes classify the same claim as a numeric target and sometimes as a structural target.
This variation does not contradict paper-level completion.
It shows that two completed workspaces can support the same aligned paper claim while recording different kinds of target-level evidence as sufficient.
We therefore report judgment variation separately from paper-anchored numeric fidelity.

\section{Discussion}\label{sec:discussion}
The case studies show that Paper-replication can bring the papers in this corpus to a completed workspace state.
All twelve runs pass the completion gate, every recorded target reaches \(\mathrm{MATCHED}\), and every matched target has report coverage.
This result should be read with the definition of completion used here.
Completion does not mean that every possible aspect of the paper has been replicated.
It is the workspace state defined by Eq.~\eqref{eq:completion_gate}: every recorded target is matched, no target remains active, the report exists, and the external checks accept the specification, progress, provenance, comparison evidence, and report coverage.
Completion therefore remains relative to the target set recorded in the reproduction matrix and to the validation checks applied to that target set.

The persistent workspace makes this completed state inspectable.
The reproduction matrix fixes the target list, the task ledger records the active target and unresolved work, and the run and provenance records link each accepted output to the command, code, configuration, seed, and paper passages used to justify the implementation.
These records let the agent resume after interruptions without relying on the chat transcript.
They also let the analysis separate completion from the path taken to reach completion.
The same completed corpus contains stable decompositions for PINN-I and SINDy, variable decompositions for PIFT and PINN-II, and different amounts of correction work before final evidence is accepted.
These differences remain visible because Paper-replication treats the workspace, not the chat transcript, as the record of the run.

The validation checks change what can count as accepted evidence.
A generated figure, table, or scalar does not satisfy a target unless the workspace links it to a successful run, method provenance, comparison evidence, and report coverage.
The paper-asset checks reduce the risk that copied paper material counts as reproduced output, and the provenance checks reduce the risk that a substitute method is accepted only because it gives a similar-looking result.
These checks do not prove that the agent's reconstruction is the only implementation consistent with the paper.
They make the evidentiary standard explicit: accepted evidence must be generated inside the replication workspace, tied to a recorded reconstruction of the paper's method, judged under the target's recorded acceptance rule, and included in the final report.

The results of the correction work show why this recording matters for scientific machine learning papers.
The agent often has to test assumptions about preprocessing, optimization, sampling, and plotting before it reaches an implementation that satisfies the target-level checks.
Superseded tracked executions do not mean that the workspace failed.
They are part of the recorded trial-and-correction path rather than work overwritten by the final output.
This is especially visible in the PINN papers, where repeated training and optimization choices account for much of both elapsed replication time and correction work.
The final accepted outputs are therefore not standalone artifacts.
They come at the end of a recorded sequence of hypotheses, runs, rejected choices, and checks.

The distinction between workspace matching and paper-anchored numeric fidelity is central to the analysis.
Paper claims are scalar, distributional, structural, and visual, and one comparison rule does not cover all of them.
A target-level match reflects the acceptance rule recorded for that target, whereas the paper-anchored scalar analysis applies a fixed threshold from the source paper's reported accuracy scale.
This separation allows the analysis to inspect both what the workspace accepted as sufficient evidence and whether the reproduced scalar quantities fall within the paper's stated accuracy class.
It also prevents the \(\mathrm{MATCHED}\) label from being treated as a single universal metric of fidelity~\citep{manheim2018categorizing,weng2024rewardhacking}.

The case studies also show that source papers differ in how fully they specify the inputs needed for replication.
PIFT and several SINDy examples specify governing equations and observation or noise rules, so the workspaces regenerate inputs and record choices.
The PINN papers point to public \texttt{.mat} fields and implicit Runge--Kutta tables, but the runs still have to generate training subsets and noisy variants because the papers do not publish sampled points or random seeds.
SINDy's Hopf example prints the normal-form equations but not the noise magnitude, sampled parameter values, or total-variation derivative settings.
The cylinder-wake example leaves a different gap: the paper materials describe direct numerical simulation followed by proper orthogonal decomposition, but they do not include the simulation trajectories, reduced-order trajectories, or preprocessing conventions needed to verify digit-level coefficient equality.
Paper-replication makes these gaps visible by documenting assumptions, deviations, and the acceptance rules used to evaluate the resulting evidence.

The repeated-run design also affects how the statistical summaries should be interpreted.
In stochastic training settings, a run is a single draw from the training procedure rather than a fixed measurement~\citep{bouthillier2021accounting}.
The hierarchical models summarize run-to-run movement and paper-level differences within this corpus, but four papers do not tightly identify between-paper variation.
For scalar anchors, PINN-II has the largest run-residual scale, but the between-paper components remain wide.
We therefore use the models to describe the observed corpus rather than to rank paper difficulty.
The effort results require the same caution: the longer PINN runs show repeated optimization and correction work in these workspaces, not that the PINN papers are intrinsically harder to replicate.

This study has limitations.
It uses four scientific machine learning papers, three runs per paper, and one coding-agent interface, model, and reasoning setting.
It does not include an ablation without the skill, so it characterizes what Paper-replication produces rather than estimating the workflow's effect on an unstructured prompt.
The repeated runs support run-level comparisons, but the corpus remains too small to draw strong conclusions about between-paper variation.
The paper-anchored fidelity analysis also uses only thirteen scalar anchors.
We evaluate PIFT using distributional and structural evidence because its paper-level claims do not provide a scalar anchor that is consistent across runs.
Each distributional or structural target has one recorded comparison per run, so repeated independent judging of the same evidence remains future work.
Finally, the scalar thresholds are derived from the accuracy scales reported in the source papers.
A different defensible threshold could change how many scalar anchors fall inside the paper-anchored threshold, but it would not change the observed run-to-run differences in the reproduced values.

\section{Conclusion}\label{sec:conclusion}
Paper-replication turns paper replication into a target-level evidence task for a coding agent.
Starting from paper materials, the workflow has the agent build a replication workspace, record the target set in the reproduction matrix, reconstruct the paper's method in specification files, and link each matched target to a generated output, a successful run, a provenance record, an acceptance rule, comparison evidence, and report coverage.
Two mechanisms support this process: persistent workspace records and validation checks outside the agent.
Together, they make completion a workspace state rather than a statement in the agent's final message.

The case studies show that Paper-replication can bring the papers in this corpus to a completed workspace state.
Across twelve independent runs on four scientific machine learning papers, every recorded target reaches \(\mathrm{MATCHED}\), no target remains active, the final replication report exists, and the completion gate accepts the specification, progress, provenance, comparison evidence, and report coverage.
This result should be read in light of the definition of completion used here.
It does not mean that every possible aspect of each paper has been replicated.
It means the recorded target set is complete under the validation checks applied to it.

The workspaces also preserve variation that a completion label alone would hide.
Repeated runs differ in target decomposition, paper-anchored scalar fidelity, elapsed replication time, correction work, and recorded acceptance rule type.
These records show that a target can be matched under its acceptance rule, even as fidelity, effort, and judgment still vary across runs.
The case studies, therefore, show why paper replication should be evaluated by the recorded process, not solely by exact numerical agreement.

Paper-replication shows how harness engineering can support coding agents in paper replication by decomposing the task into targets and evidence records, linking those records to provenance, comparisons, and report coverage, and applying validation checks before completion.
By storing workflow state in workspace files and requiring validation checks before a target can count as matched, the workflow makes the agent's reconstruction, generated outputs, accepted evidence, and deviations available for inspection.
Paper-replication is therefore an inspectable replication process, not a guarantee of exact reproduction or proof that a completed workspace covers every possible paper claim.

\section*{Code and data availability}
We release Paper-replication as Codex and Claude Code coding-agent skills, along with the initial and follow-up prompts given to the Codex agent, twelve agent-generated case-study workspaces, and the analysis files and scripts used to produce all results in this manuscript: \url{https://github.com/PredictiveScienceLab/paper-replication-paper}.

\section*{CRediT authorship contribution statement}
\textbf{Atharva Hans:} Conceptualization, Methodology, Software, Validation, Formal analysis, Investigation, Visualization, Writing -- original draft, Writing -- review \& editing.
\textbf{Ilias Bilionis:} Conceptualization, Methodology, Supervision, Funding acquisition, Writing -- review \& editing.

\section*{Declaration of competing interest}
The authors declare no competing interests.

\section*{Declaration of generative AI and AI-assisted technologies in the manuscript preparation process}
A Codex coding agent equipped with the paper-replication skill was used to produce the case-study reproductions analyzed in this work.
ChatGPT was used to assist with manuscript editing and writing.
The authors reviewed and edited all generated material carefully and take responsibility for the final content of the manuscript.

\section*{Acknowledgements}
The authors thank Eli Lilly and Company for supporting this work.

\bibliographystyle{elsarticle-harv}
\bibliography{bibliography}

\end{document}